\documentclass[11pt,a4paper]{llncs}
\usepackage{amsmath}
\usepackage{amssymb}
\usepackage{graphicx}
\usepackage{marvosym}
\usepackage{url}
\usepackage{fancyhdr}
\usepackage{wrapfig}
\usepackage{geometry}
\usepackage{graphicx,lipsum}
\usepackage[export]{adjustbox}
\newcommand{\keywords}[1]{\par\addvspace\baselineskip
\noindent\keywordname\enspace\ignorespaces#1}

\fancypagestyle{firstpage}{\fancyhf{}
}

\begin{document}


\title{\LARGE{ChatGPT Alternative Solutions: Large Language Models Survey}}


%
%
\author{\large{Hanieh Alipour \and Nick Pendar \and Kohinoor Roy}}
\institute{\large{SAP Company Inc.\\
\large{hanieh.alipour@sap.com,nick.pendar@sap.com,kohinoor.roy@sap.com}}}

\maketitle

\thispagestyle{firstpage}

\begin{abstract}
In recent times, the grandeur of Large Language Models (LLMs) has not only shone in the realm of natural language processing but has also cast its brilliance across a vast array of applications. This remarkable display of LLM capabilities has ignited a surge in research contributions within this domain, spanning a diverse spectrum of topics. These contributions encompass advancements in neural network architecture, context length enhancements, model alignment, training datasets, benchmarking, efficiency improvements, and more.
Recent years have witnessed a dynamic synergy between academia and industry, propelling the field of LLM research to new heights. A notable milestone in this journey is the introduction of ChatGPT, a powerful AI chatbot grounded in LLMs, which has garnered widespread societal attention. The evolving technology of LLMs has begun to reshape the landscape of the entire AI community, promising a revolutionary shift in the way we create and employ AI algorithms.
Given this swift-paced technical evolution, our survey embarks on a journey to encapsulate the recent strides made in the world of LLMs. Through an exploration of the background, key discoveries, and prevailing methodologies, we offer an up-to-the-minute review of the literature. By examining multiple LLM models, our paper not only presents a comprehensive overview but also charts a course that identifies existing challenges and points toward potential future research trajectories. This survey furnishes a well-rounded perspective on the current state of generative AI, shedding light on opportunities for further exploration, enhancement, and innovation.
\end{abstract}

\section{Introduction}

Language is a fundamental aspect of human cognition, enabling expression and communication. Machines, however, lack an inherent ability to understand and use human language unless equipped with robust AI algorithms. Bridging this gap to make machines communicate like humans has been a significant research challenge. Bridging this divide has long been a formidable research challenge, to endow machines with the ability to read, write, and communicate in a manner akin to humans \cite{y2022large} \cite{hauser2002faculty}. Language modeling seeks to construct models that estimate the generative likelihood of word sequences, thereby facilitating the prediction of probabilities associated with forthcoming or missing tokens \cite{rosenfeld2000two}\cite{wang2019superglue}.

Recent times have borne witness to significant breakthroughs in the realm of language models, largely attributed to advancements in deep learning techniques, innovations in neural architectures such as transformers, enhanced computational capabilities, and the widespread availability of vast training datasets extracted from the internet. These collective advancements have ushered in a trans-formative era, empowering the creation of LLMs that approach or even approximate human-level performance on select evaluation benchmarks \cite{adiwardana2020towards}.

Notwithstanding the considerable advancements and impact achieved in the realm of LLMs, the foundational principles underpinning their functioning remain relatively under-explored. The task of training proficient LLMs presents a formidable challenge for the research community. Given the immense computational resources required, conducting repetitive, exhaustive investigations into the effects of various training strategies for LLMs becomes prohibitively costly. It is worth noting that the primary custodians of LLM training are typically entities in the industry, and many critical aspects of this process, such as data collection and cleansing methods, remain undisclosed to the public. Simultaneously, the endeavor to align LLMs with human values and preferences poses a substantial challenge \cite{GPT4}. However, despite the aforementioned challenges, there is an ongoing and pressing need for the utilization and exploration of LLMs. Hence, the motivation behind this paper is to provide an extensive review of popular LLMs, both open-source and proprietary, to assist individuals in navigating the landscape and harnessing the potential of these models.
\begin{figure*}[!h]
  \centering
   \includegraphics[width=\linewidth,height=5.5cm]{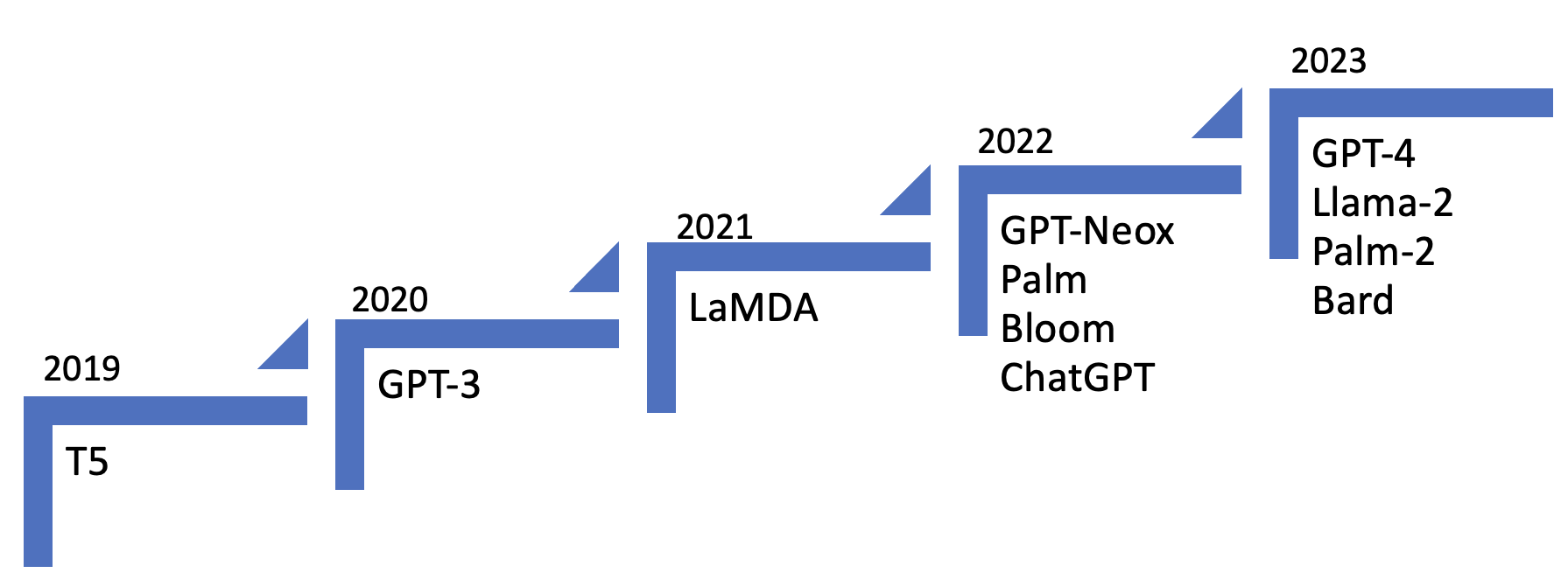}
  \caption{Chronology of existing Large Language Models.}
  \label{fig:timeline}
 \end{figure*}

In this paper, we conduct a comprehensive review of various LLM solutions, both open-source and close-sourced. Figure \ref{fig:timeline} showcases the names of noteworthy releases of LLMs. A detailed comparison of the features of these services is provided to highlight their respective strengths and capabilities. By examining these offerings, we aim to provide valuable insights into the landscape of LLMs, aiding researchers, developers, and the broader AI community in making informed decisions regarding their application and utilization. In addition to discussing the key limitations of LLMs like ChatGPT, we thoroughly evaluate both their strengths and the areas that require further enhancements. The subsequent sections of this survey are structured as follows: Section 2 provides an introduction to the background of LLMs, followed by an exposition on prompt engineering for LLM development in Section 3. Then, Section 4 conducts a review of solutions for LLMs proposed by OpenAI. In Section 5, we undertake a comprehensive review and summarization of prominent alternative solutions to ChatGPT, including Bard, PaLM, T5, and LLaMA. Additionally, Section 6 offers insights into open problems and outlines potential future directions for the development of LLMs. The survey is concluded in Section 7.

\section{Literature Review}
Table \ref{table:Related-Work}  delves into the existing body of research within the field of interest. It provides an overview of key studies and contributions that have paved the way for the current state of the subject matter. 

\begin{table}[htb]
\caption{Summary of Related Work}\label{table:Related-Work}
\begin{center}
\begin{tabular}{|p{3cm} | p{2cm}|p{10cm}| }
\hline
 Citation and Year & Methodology & Focuses \\
\hline
\cite{kasneci2023chatgpt} 2023  & Survey & They present the potential benefits and challenges of educational applications of large language models, from student and teacher perspectives. We briefly discuss the current state of large language models and their applications. We then highlight how these models can be used to create educational content, improve student engagement and interaction, and personalize learning experiences.  \\
\hline
\cite{hadi2023survey} 2023 & Survey & They provide an overview of the history and evolution of large language models (LLMs) along with various training methods. It explores their diverse applications in fields like medicine, education, finance, and engineering. The paper also discusses how LLMs are influencing the future of AI and their potential for solving real-world problems.\\
\hline
\cite{min2023recent} 2021 & Survey &They present instances wherein large language models are employed to address Natural Language Processing (NLP) tasks through techniques such as pre-training followed by fine-tuning, prompting, or text generation methodologies. Furthermore, they introduce strategies utilizing pre-trained language models to generate supplementary data for purposes such as training augmentation.  \\
\hline
\cite{liu2023summary} 2023 & Survey & They offer an extensive examination of research related to ChatGPT, the latest advancements in LLMs from the GPT series, and their potential applications across a wide range of domains. The authors conducted a thorough analysis of 194 pertinent papers on arXiv, including trend analysis, word cloud visualization, and distribution analysis across various application domains. \\
\hline

\cite{zhu2023large} 2023 & Survey & They have undertaken an extensive investigation into the profound influence of LLMs on Information Retrieval across multiple facets. They have categorized existing methods into specific functional groups, which include query rewriting, retrieval, re-ranking, and reader modules. Within the domain of query rewriting, LLMs have proven their efficacy in comprehending ambiguous or multifaceted queries, thereby improving the precision of intent identification.\\

\hline
\end{tabular}

\end{center}
\end{table}

\section{Prompt Engineering}

Prompt engineering refers to the deliberate design and construction of prompts or inputs given to a language model or an AI system. In the context of natural language processing (NLP) models like GPT (Generative Pre-trained Transformer), prompt engineering involves crafting input queries or instructions in a way that elicits desired and contextually appropriate responses from the model. The goal of prompt engineering is to improve the performance and output of the AI system by guiding it to generate responses that align with the user's expectations or specific objectives \cite{zhou2022large}. 

\subsection{Zero-shot and Few-shot learning}

The fundamental techniques for prompting the model, namely zero-shot and few-shot learning, have been pioneered in numerous LLM papers and are frequently employed for benchmarking LLM performance cite{zhao2021calibrate}.

\textbf{Zero-shot learning} is to simply feed the task text to the model and ask for results.

\textbf{Few-shot learning} involves providing the model with a collection of exemplary demonstrations, each comprising both input and desired output, related to the target task. By exposing the model to these illustrative examples initially, it gains a deeper understanding of human intentions and the criteria for generating desired responses. Consequently, few-shot learning often results in enhanced performance compared to zero-shot learning. However, this advantage comes with the trade-off of increased token consumption and the potential challenge of reaching the context length limit, particularly when dealing with lengthy input and output text.

\subsection{Chain-of-Thought} 

One of the well-known and easily applicable prompt engineering techniques involves adding the phrase "Think step by step" to the end of a prompt. Researchers from the University of Tokyo and Google conducted a study revealing that the inclusion of this expression significantly improved the accuracy of GPT-3 (text-davinci-002 model) across various tasks. For example, it elevated accuracy from 17.7\% to 78.7\% on the MultiArith test \cite{kojima2022large}, which involves arithmetic questions requiring multiple steps for resolution. Prystawski and colleagues provide insights into the effectiveness of "think step by step" and suggest reasons for its efficacy \cite{prystawski2023think}. It is worth noting that this approach has been observed to be less beneficial (adding less value) on more advanced GPT models like GPT-4.

In this paper\cite{wei2022chain}, The researchers investigated the utility of chain-of-thought prompting as a straightforward and widely applicable technique to augment reasoning capabilities in language models. By conducting experiments involving arithmetic, symbolic, and commonsense reasoning, they observed that chain-of-thought reasoning emerges as a property of model scale, enabling sufficiently LLMs to excel in reasoning tasks that exhibit flat scaling curves under alternative conditions.

\begin{figure}[!h]
  \centering
   \includegraphics[width=\linewidth]{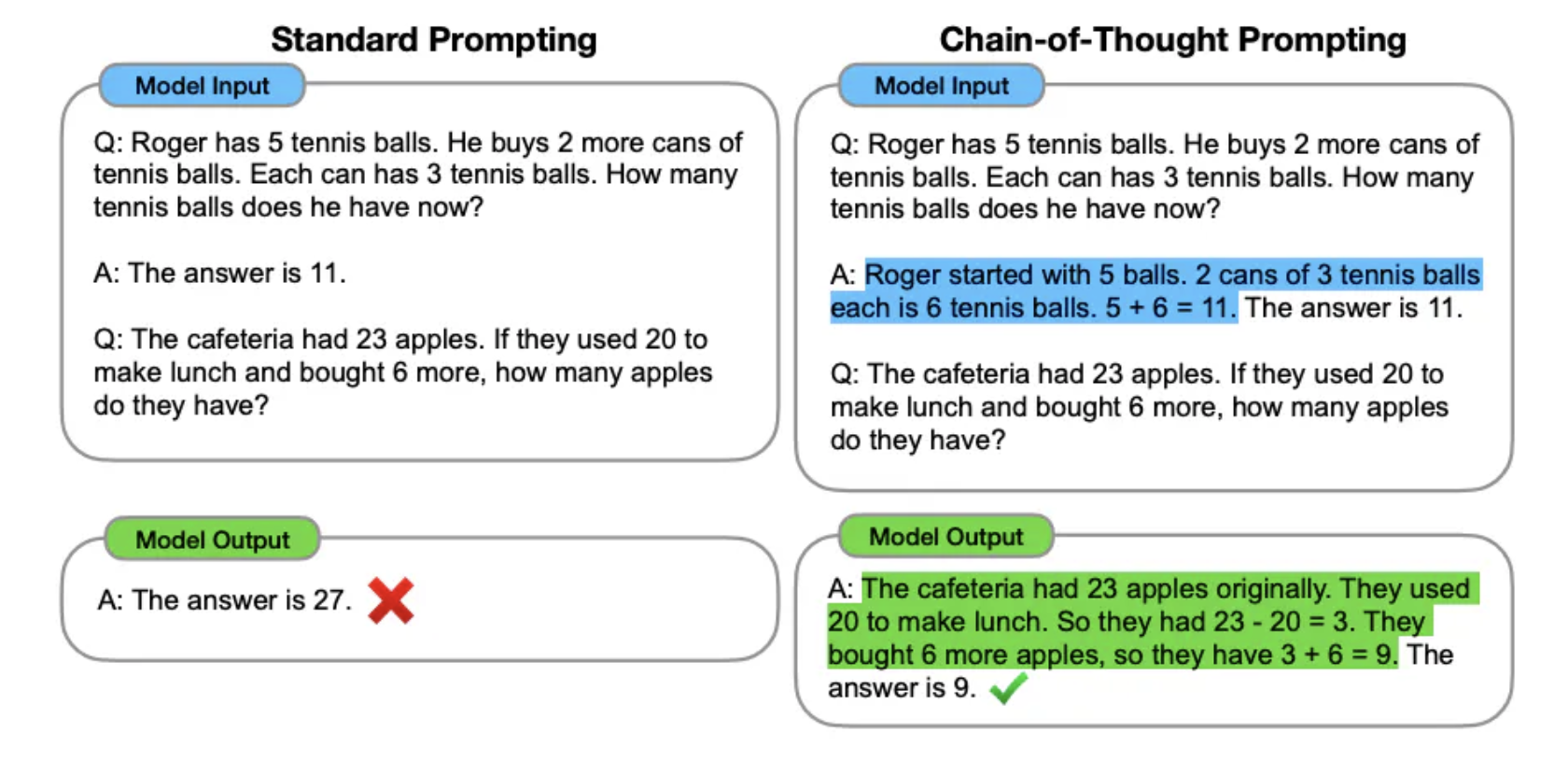}
  \caption{Chain-of-thought prompting example \cite{wei2022chain}}
  \label{fig:COT}
 \end{figure}

\section{OpenAI}

OpenAI, established in December 2015, is both an artificial intelligence research laboratory and a technology company dedicated to the advancement and widespread adoption of artificial general intelligence (AGI). Their primary objective is to ensure that AGI, referring to highly autonomous systems surpassing human performance in economically valuable tasks, is harnessed for the collective benefit of humanity \cite{ray2023chatgpt}.
The name "OpenAI" epitomizes their commitment to transparency and openness in their pursuits. Positioned at the forefront of AI research, OpenAI has made significant contributions to the field, notably the development of the GPT (Generative Pre-trained Transformer) series.
Through a multifaceted approach, OpenAI engages in extensive research and innovation across diverse domains such as natural language processing, robotics, computer vision, and reinforcement learning. In the spirit of collaboration and knowledge dissemination, they have also made available various models and tools to the public and research community, fostering progress and understanding in the AI landscape \cite{openAI}.

\subsection{ChatGPT}
ChatGPT is an AI-driven conversational agent that has been developed by OpenAI. It is built upon the advanced GPT (Generative Pre-trained Transformer) architecture, which is widely recognized as a state-of-the-art language model. The primary objective of ChatGPT is to generate responses in natural language that closely resemble human-like interactions. Through its training on extensive amounts of internet text data, ChatGPT has acquired the ability to comprehend user input and produce coherent and contextually appropriate responses \cite{gill2023chatgpt}.

ChatGPT has garnered attention for its versatility, excelling in engaging users across various conversational topics and delivering informative and creative responses. It finds application in tasks like text completion, question-answering, and interactive dialogue. However, it's important to recognize that despite its impressive performance, ChatGPT can occasionally produce incorrect or nonsensical responses, and its sensitivity to input phrasing and context should be taken into account. OpenAI has introduced multiple versions of ChatGPT, consistently refining and enhancing its functionality over time. It has been made accessible to the public through various platforms and APIs, facilitating integration by developers and users into their own applications and systems. The functioning of GPT-3.5 involves a structured process comprising three distinct steps (Figure \ref{fig:gpt3}):

\begin{figure*}
    \centering
    \includegraphics[width=\textwidth,height=8cm]{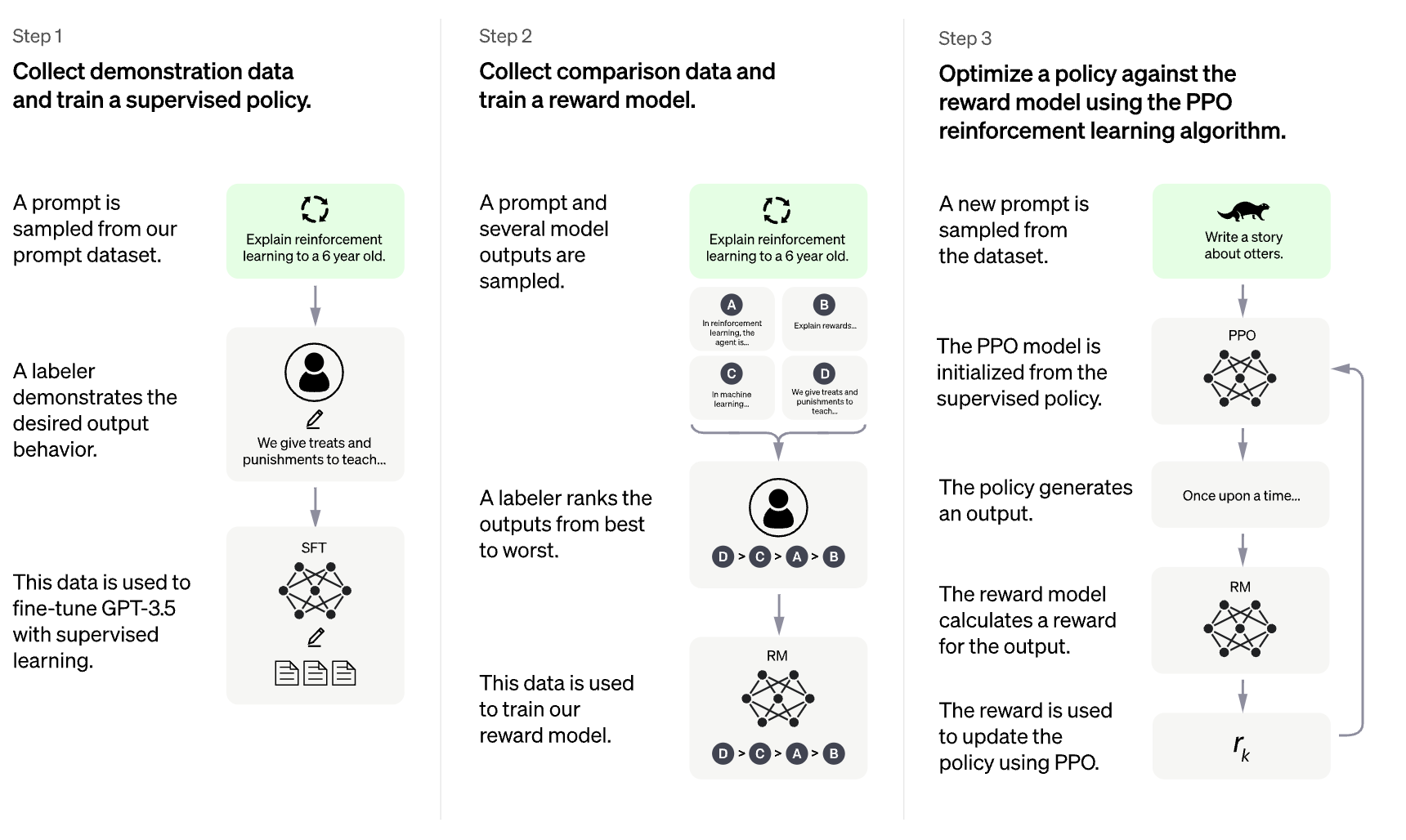}
    \caption{
        GPT-3.5 model workflow \cite{ray2023chatgpt}
    }
    \label{fig:gpt3}
\end{figure*}

\begin{itemize}

  \item The process involves obtaining demonstration data where a randomly selected prompt is labeled with the desired output behavior. This data is then used to fine-tune GPT-3 using supervised learning principles.
  \item  The process involves selecting a prompt and generating multiple model-generated outputs. A labeler ranks these outputs based on quality, from best to worst. This comparison data is used to train the reward model, helping the system learn how to assess the quality of its generated responses.
  \item  In the final step, a new prompt is chosen, and the policy generates an output. A reward model assesses the generated output and provides a reward value. This reward value is used to update the policy using the proximal policy optimization (PPO) algorithm, improving the system's responses gradually to make them more desirable and accurate.

\end{itemize}

OpenAI \cite{openAI} describes GPT-4 as: "10 times more advanced than its predecessor, GPT-3.5. This enhancement enables the model to better understand the context and distinguish nuances, resulting in more accurate and coherent responses." The unpaid version of ChatGPT lacks internet access, while ChatGPT Plus, based on the GPT-4 architecture, has this capability. Recently, OpenAI incorporated a 'Browse with Bing' feature into GPT-4, enabling the AI chatbot to browse the internet and furnish information on contemporary events. A notable distinction between Chat GPT-4 and Chat GPT-3 is the significant enhancement in the former's capacity to comprehend images. This augmentation stems from the multimodal nature of Chat GPT-4, denoting its capability to interpret diverse modes of information, encompassing both textual and visual elements. In contrast, the preceding version, Chat GPT-3, was confined to processing and generating responses based solely on textual inputs, thereby constraining its applicability across various use cases \cite{openAI}.

Key Features of ChatGPT-4 \cite{koubaa2023gpt,chatgpt4}:
\begin{itemize}
\item	Expanded Context Window: A distinctive attribute of ChatGPT-4 lies in its heightened capacity to consider a broader context when processing language. The enlarged context window empowers the model to grasp intricate linguistic patterns more effectively, yielding responses that exhibit greater accuracy and coherence.
\item	Conversational Ability: ChatGPT-4 demonstrates improved conversational prowess, contributing to interactions with the model that feel more natural and engaging. This advancement allows the model to engage in longer and contextually coherent dialogues with users, resulting in the exchange of more relevant and meaningful information.
\item	Diverse Use Cases: ChatGPT-4 proves versatile across various industries, excelling in applications such as virtual assistants, text summarization, content creation, and beyond. Its adaptability renders it valuable in a multitude of real-world scenarios, showcasing its efficacy in addressing diverse needs and requirements.
\end{itemize}

\subsection{OpenAI Playground}

OpenAI Playground and ChatGPT share significant similarities as they originate from the same AI research organization. Both platforms employ comparable generative AI models, but their primary distinction lies in their intended users. While ChatGPT targets the public, OpenAI Playground caters more to developers seeking to explore OpenAI's diverse AI offerings \cite{openAI}. Table \ref{table:Playground} presents the objective of OpenAI Playground and ChatGPT.

\begin{table}
\begin{center}
\caption{Purposes of OpenAI Playground and ChatGPT}\label{table:Playground}
\begin{tabular}{|p{7cm} | p{7cm}|}
\hline
OpenAI Playground & ChatGPT 3.5  \\
\hline
Enables users to conduct experiments with diverse machine learning models  & Engaging in interactive conversations with the user   \\
\hline
Can be fine-tuned with custom data sets & Generating responses to user prompts using natural language \\
\hline
Valuable for creating applications dependent on machine learning & Executing writing tasks, such as generating text passages   \\
\hline
In addition to other models, it grants access to GPT & Conducting translation tasks efficiently \\
\hline
\end{tabular}

\end{center}
\end{table}

The OpenAI Playground offers real-time interactivity with AI models, enabling users to input text and receive instant responses. The main features are:

\begin{itemize}

\item  Diversity of Models: The platform boasts an array of AI models, each showcasing unique capabilities. From language models like GPT-4 to image models like DALL-E, users can access a variety of cutting-edge technologies.
\item  User-Friendly Interface: OpenAI Playground is thoughtfully designed with simplicity and intuitiveness in mind. Even individuals with limited AI experience can effortlessly navigate and utilize the platform.
\item Educational Value: Beyond being a mere tool, the OpenAI Playground serves as an invaluable learning resource. It empowers users to grasp the fundamentals of AI and gain insights into the evolving landscape of AI technology.

\end{itemize}

\section{Alternative Solutions}

\subsection{OpenAssistance}

OpenAssistance is a project aimed at developing a chatbot that utilizes LLM and employs Reinforcement Learning with Human Feedback (RLHF) techniques. The primary objective of the project is to enhance the chatbot's ability to answer questions while effectively following instructions. RLHF involves training the chatbot by providing it with feedback from human evaluators, allowing it to learn and improve its responses over time. By combining the power of LLM and reinforcement learning, OpenAssistance seeks to create a more adaptive and intelligent chatbot that can provide accurate and helpful answers to a wide range of inquiries. Given that the models are predominantly licensed under the Apache 2.0 license, it is reasonable to infer that commercial utilization of models derived from them as closed source is prohibited. However, the license permits unrestricted usage, modification, and distribution of the models in open-source scenarios. By the Apache 2.0 license, any modified derivatives must indicate their modifications prominently and preserve all original copyright notices. The conversation tree is the fundamental data structure in OpenAssistant.


The researchers \cite{mu2023navigating} evaluated the zero-shot performance of ChatGPT and OpenAssistant within the domain of Computational Social Science classification tasks. Additionally, they examined the effects of different prompting strategies. Their experiment encompassed several factors, such as prompt complexity. This involved investigating the outcomes of incorporating label definitions into the prompt, utilizing synonyms for label names, and exploring the influence of integrating memories during the training of the foundation model. 

\begin{table*}[h]
\centering
\caption{LLMs zero-shot classification results across all prompt settings. All datasets are evaluated with accuracy and macro-F1 scores \cite{mu2023navigating}.}
\label{tab:OP} 
\begin{tabular}{|l|l|l|l|l|l|l|}
\hline
Model & \multicolumn{2}{c|}{Complaint} & \multicolumn{2}{c|}{Vaccine Stance} & \multicolumn{2}{c|}{Bragging} \\
  
  & Accuracy & F1-macro & Accuracy & F1-macro  & Accuracy & F1-macro \\
  \hline
  Logistic Regression & 81.4 & 79.7 & 72.8 &73.1 & 88.6 & 58.8 \\
  \hline

  BERT-large & 89.4 & 88.6 & 81.5 & 81.3 & 91.3 & 76.1 \\
\hline
GPT Basic & 89.7 & 88.7 & 73.0 & 73.8 & 85.1 & 67.6  \\
\hline
GPT T/L Desc & 89.0 & 88.0 & 73.3 & 73.7 & 84.9 & 67.4 \\
\hline
GPT Memory Recall & 87.1 & 86.4 & 66.2 & 66.9 & -& - \\
\hline
OA Basic & 72.3 & 72.3 & 61.7 & 60.3 & 89.3 & 57.6 \\
\hline
OA T/L Desc & 65.3 & 65.2 & 73.7 & 73.6 & 88.4 & 48.2 \\
\hline
OA Memory Recall & 82.6 & 82.1 & 64.2 & 63.8 & - & -  \\
\hline
 Model & \multicolumn{2}{c|}{Rumor Stance} & \multicolumn{2}{c|}{Sarcasm} & \multicolumn{2}{c|}{Hata Speech} \\

  & Accuracy & F1-macro & Accuracy & F1-macro  & Accuracy & F1-macro \\
  \hline
  Logistic Regression & 68.5 & 40.9 & 76.1 & 53.5 & 83.2 & 79.2 \\
\hline
BERT-large & 73.2 & 48.2 & 78.9 & 58.4 & 84.5 & 81.2 \\
\hline
GPT Basic & 49.4 & 33.4 & 67.3 & 62.1 & 75.5 & 72.4 \\
\hline
GPT T/L Desc & 59.2 & 45.7 & 61.3 & 57.9 & 76.9 & 72.1 \\
\hline
GPT Memory Recall & 40.2 & 30.9 & - & -& 71.7 & 69.6 \\
\hline
OA Basic & 45.2 & 27.4 & 71.9 & 48.6 & 63.5 & 63.3 \\
\hline
OA T/L Desc & 56.2 & 29.0 & 75.9 & 49.9 & 75.5 & 73.3 \\
\hline
OA Memory Recall & 52.4 & 34.6 & -& -& 55.4 & 55.4 \\
\hline 
\end{tabular}
\end{table*}

In the paper \cite{mu2023navigating}, Table \ref{tab:OP}, the prediction results of all zero-shot LLMs for each round are depicted. The researchers made several noteworthy observations. Firstly, they noted that supervised baselines generally outperformed LLMs across most prompt settings in most tasks (4 out of 6). Furthermore, when considering the F1-macro measure, GPT consistently exhibited superior performance compared to OpenAssistant across all prompt settings and tasks. However, their findings indicated that OpenAssistant achieved higher accuracy than GPT in certain imbalanced datasets, such as the 'Bragging and Sarcasm' task. This disparity may arise from OpenAssistant defaulting to the neutral class, encompassing labels that lack any specific speech act, such as 'Not Bragging and Not Sarcastic'. The findings from the study suggest that in a zero-shot scenario, the existing LLMs fail to achieve the performance levels of smaller baseline transformer models that have undergone fine-tuning, such as BERT. Furthermore, the researchers discovered that different prompting strategies have a substantial impact on classification accuracy, with variations in both accuracy and F1 scores exceeding 10\%. Notably, GPT demonstrates the highest predictive performance in two specific downstream tasks related to speech act detection, specifically the Complaint task with an accuracy of 89.7 and an F1-macro score of 88.7.

According to \cite{kopf2023openassistant}, OpenAssistant can compete with GPT-3.5 Turbo, the direct predecessor of GPT-4. Also, they claimed human users prefer their assistant sitting on the fine-tuned Pythia-12B to the answers generated by GPT-3.5 Turbo by 48.3 percent. A scientific evaluation of the capabilities of the model is currently not available, the evaluation in the paper refers to human preferences alone.

\subsection{LLaMA}

Meta, the parent company of Facebook, has made a notable announcement by unveiling their latest breakthrough in the field of artificial intelligence (AI): LLaMA (Large Language Model Meta AI), a cutting-edge LLM. Meta AI embarked on a mission to develop a series of Long Language Models (LLMs) that excel in different inference scenarios. The outcome of their endeavor is the LLaMA collection. These models are designed to be smaller than existing LLMs, yet they undergo training with a larger number of tokens. This approach enhances their performance and simplifies the process of retraining and fine-tuning the models for specific real-world applications.

The LLaMA models are constructed using a transformer architecture \cite{vaswani2017attention}, incorporating various enhancements inspired by other models. To improve training stability, the LLaMA models employ the RMSNorm normalizing function \cite{zhang2019root}, initially introduced by GPT-3. Additionally, they replace the ReLU non-linearity with the SwiGLU activation function from PaLM \cite{shazeer2020glu}, resulting in improved model performance. To harness positional information more effectively, the LLaMA models utilize rotary positional embeddings (RoPE) from GPTNeo \cite{su2021roformer}, which offer advantages over absolute positional embeddings.  

LLaMA models come in four different sizes: 7, 13, 33, and 65 billion parameters \cite{touvron2302llama}. It operates by accepting a sequence of words as input and recursively generates text by predicting the next word. To train the model, they selected text samples from the top 20 languages spoken worldwide, with particular emphasis on languages that utilize the Latin and Cyrillic alphabets. Also, they use only publicly available data, for example, English CommonCrawl, C4, Github, Wikipedia, Gutenberg and Books3, ArXiv, and Stack Exchange. 

The original LLaMA code is GPL licensed which means any project using it must also be released under GPL. LLaMA has GNU general public license, however, LLaMA models are licensed for research use only, which prevents commercial use of those models.

In this paper \cite{touvron2302llama}, they compare LLaMA with other foundation models, namely the non-publicly available language models GPT-3 \cite{brown2020language}, Gopher \cite{rae2021scaling}, Chinchilla \cite{hoffmann2022training} and PaLM \cite{chowdhery2022palm}, as well as the open-sourced OPT models \cite{zhang2022opt}, GPT-J \cite{wang2021gpt}, and GPTNeo \cite{black2022gpt}. 



The results of their evaluation for different use cases are \cite{touvron2302llama}:

\begin{itemize}

\item \textit{For the Common Sense Reasoning use case}, LLaMA-65B outperforms Chinchilla-70B on all reported benchmarks but BoolQ. Similarly, this model surpasses PalM-540B everywhere but on BoolQ and WinoGrande. LLaMA-13B model also outperforms GPT-3 on most benchmarks despite being 10× smaller.

\item \textit{For Closed-book Question Answering use case}, On both, LLaMA-65B achieves state-of-the-art performance in the zero-shot and few-shot settings. More importantly, the LLaMA-13B is also competitive on these benchmarks with GPT-3 and Chinchilla, despite being 5–10× smaller. This model runs on a single V100 GPU during inference.

\item \textit{For Reading Comprehension use case}, On these benchmarks, LLaMA-65B is competitive with PaLM-540B, and LLaMA-13B outperforms GPT-3 by a few percent.

\item \textit{For Mathematical Reasoning use case}, On GSM8k, LLaMA-65B outperforms Minerva-62B, although it has not been fine-tuned on mathematical data.

\item \textit{For Code Generation use case},  For a similar number of parameters, LLaMA outperforms other general models such as LaMDA and PaLM, which are not trained or fine-tuned specifically for code. LLaMA with 13B parameters and more outperforms LaMDA 137B on both HumanEval and MBPP. LLaMA 65B also outperforms PaLM 62B, even when it is trained longer.

\item \textit{For Massive Multitask Language Understanding (MMLU)} LLaMA-65B is behind both Chinchilla-70B and PaLM-540B by a few percent on average and across most domains. A potential explanation is that a limited number of books and academic papers are used in the pre-training data, i.e., ArXiv, Gutenberg, and Books3, which sums up to only 177GB, while these models were trained on up to 2TB of books.

\end{itemize}

Llama 2 signifies a significant advancement in the field of natural language processing, offering an open-source platform with the potential to benefit both research and commercial applications. Its versatility empowers a diverse range of users to responsibly explore and implement innovative solutions. This paper \cite{roumeliotis2023llama} not only delves into the foundational aspects of the Llama 2 model but also examines how early adopters leverage its capabilities in their AI projects. The rapid fine-tuning of the model within ten days for medical-specific domains and chatbots reflects a prevailing trend among researchers, emphasizing the pursuit of a more robust and contextually appropriate AI framework. This pursuit aims to uphold higher standards of quality and ethics in future AI applications.
Furthermore, they explore the implications of Llama 2's adoption on the broader open-source AI landscape, addressing both challenges and opportunities for developers and researchers striving to create cutting-edge AI solutions. This study \cite{roumeliotis2023llama} serves as an initial exploration of the Llama 2 pre-trained model, establishing a promising foundation for future research endeavors. By prioritizing ethical considerations, we can harness the capabilities of Llama 2 to drive positive impacts across diverse domains.

\subsection{Alpaca}

Stanford University researchers have developed an innovative natural language processing (NLP) model known as Stanford Alpaca \cite{taori2023alpaca}. This novel model has already surpassed traditional approaches in performance. Unlike current NLP models, the Stanford Alpaca NLP model focuses on generating more accurate and natural language interpretations by effectively capturing the context and interconnections among words.

Alpaca is fine-tuned based on Meta's LLaMA 7B model. The researchers trained Alpaca using 52,000 instruction-following demonstrations generated in a self-instruct style utilizing text-davinci-003. Notably, when evaluated on the self-instruct evaluation set, Alpaca demonstrates numerous behaviors akin to OpenAI's text-davinci-003 model. Additionally, Alpaca's surprising compactness and cost-effectiveness make it remarkably easy and inexpensive to reproduce \cite{taori2023alpaca}.

\begin{figure*}
    \centering
    \includegraphics[width=\textwidth,height=5.5cm]{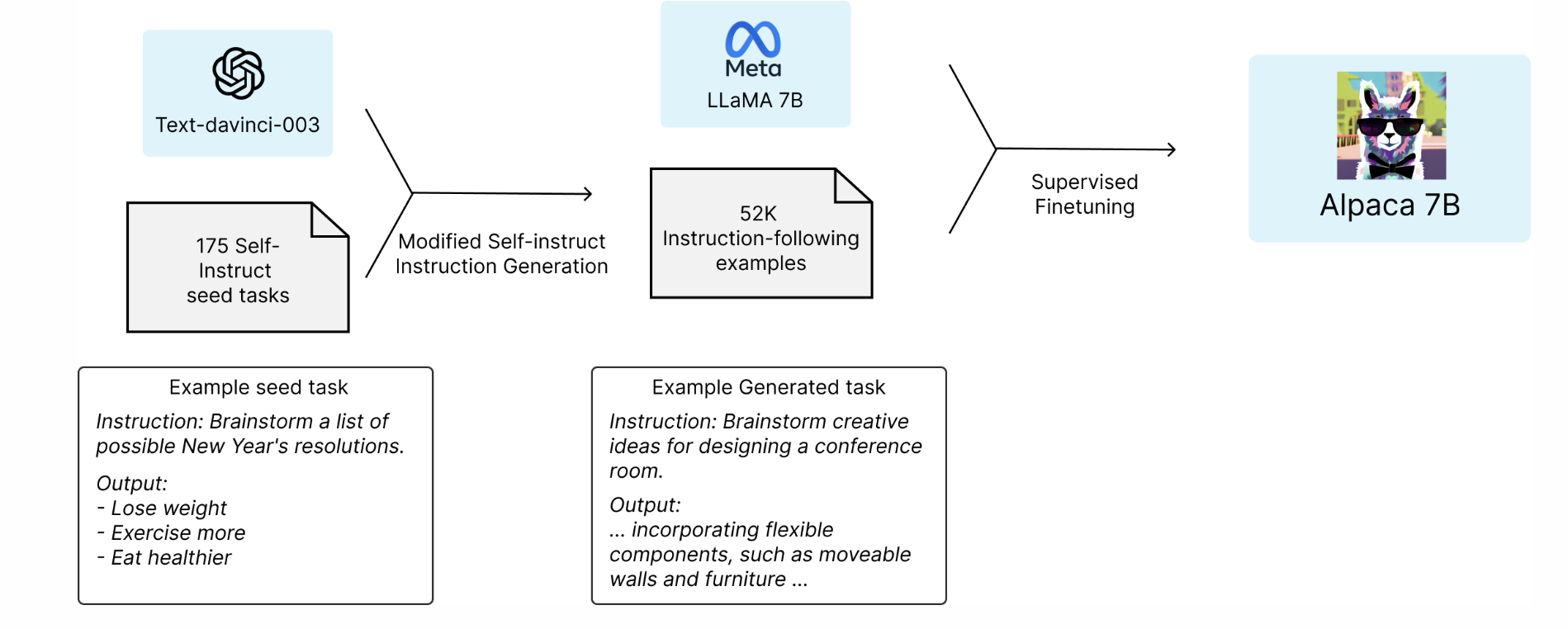}
    \caption{
        the Alpaca model \cite{taori2023alpaca}
    }
    \label{fig:Alpa}
\end{figure*}

The provided Figure \ref{fig:Alpa} outlines the methodology employed to acquire the Alpaca model. To obtain the necessary data, they generated instruction-following demonstrations by building upon the self-instruct approach. Starting with 175 human-written instruction-output pairs from the self-instruct seed set, they utilized text-davinci-003 to prompt the generation of additional instructions based on the seed set as in-context examples. This process, which involved streamlining the generation pipeline and reducing costs significantly, resulted in 52,000 unique instructions and their corresponding outputs. Notably, the entire data generation process costs less than 500 dollars using the OpenAI API.

Armed with this instruction-following dataset, they proceeded to fine-tune the LLaMA models using Hugging Face’s training framework. They leveraged techniques such as Fully Sharded Data-Parallel and mixed precision training to achieve optimal results. For their initial run, fine-tuning a 7B LLaMA model required three hours on eight 80GB A100s, costing less than 100 dollars on most cloud computing providers. It is essential to mention that training efficiency could be further improved to achieve even greater cost reduction.

They emphasized that Alpaca is strictly intended for academic research, and any commercial utilization is prohibited. Several factors contributed to this decision:
\begin{itemize}
\item Alpaca is built upon LLaMA, which has a non-commercial license, resulting in the inheritance of this restriction.
\item The instruction data used in Alpaca is derived from OpenAI's text-davinci-003, and its terms of use prohibit the development of models that compete with OpenAI.
\item Due to the lack of sufficient safety measures, Alpaca is not prepared for widespread deployment or general use.
\end{itemize}

\subsection{GPT-Neox}

\noindent GPT-NeoX, also known as GPT-NeoX-20B, stands out as a cutting-edge open-source model in the field of natural language processing (NLP). This is an autoregressive transformer decoder model with 20 billion parameters, 44 layers, a hidden dimension size of 6144, and 64 heads. It was collaboratively developed by a group of researchers associated with EleutherAI, showcasing their expertise in pushing the boundaries of NLP advancements.
GPT-NeoX-20B underwent training using the Pile dataset \cite{gao2020pile}, which is a meticulously curated and extensive dataset tailored specifically for training LLMs. The Pile consists of over 825 GB of raw text data. This dataset encompasses a wide range of information from 22 distinct data sources, categorized into five major groups for ease of organization and analysis \cite{black2022gpt}:

\begin{itemize}

\item Academic Writing: Pubmed Abstracts and PubMed Central, arXiv, FreeLaw, USPTO Backgrounds, PhilPapers, NIH Exporter.
\item Web-scrapes and Internet Resources: CommonCrawl, OpenWebText2, StackExchange, Wikipedia (English).
\item Prose: BookCorpus2, Bibliotik, Project Gutenberg.
\item Dialogue: Youtube subtitles, Ubuntu IRC, OpenSubtitles, Hacker News, EuroParl.
\item Miscellaneous: GitHub, the DeepMind Mathematics dataset, Enron Emails.

\end{itemize}

While the architecture of the model shares similarities with GPT-3, there are several noteworthy distinctions. The following are the key differences at a high level:

\begin{itemize}
\item Rotary Positional Embeddings: They used RoFormer \cite{su2021roformer} instead of the learned positional embeddings used in GPT models.

\item Parallel Attention and FF Layers: To boost efficiency in the neural network model, the Attention and Feed-Forward (FF) layers are calculated simultaneously. This approach reduces the number of all-reduce operations in both the forward and backward passes. It achieves this by employing op-sharding for each residual addition, which would typically necessitate an all-reduce operation. Calculating the Attention and Feed-Forward operations in parallel allows intermediate results to be locally reduced before a single all-reduce operation, thus minimizing communication overhead and enhancing computational efficiency during training.

\item Initialization: For the Feed-Forward output layers before the residuals, the initialization scheme introduced in \cite{Wang} is used. For all other layers, the small initialization scheme from \cite{nguyen2019transformers} is used

\item All Dense Layers: While GPT-3 uses alternating dense and sparse layers, they instead opt to exclusively use dense layers to reduce implementation complexity.
\end{itemize}

From a hardware perspective, GPT-NeoX-20B undergoes training using twelve Supermicro AS-4124GO-NART servers. Each server is equipped with eight NVIDIA A100-SXM4–40GB GPUs and configured with two AMD EPYC 7532 CPUs. The servers are interconnected using high-speed connections to facilitate efficient communication. In this work \cite{black2022gpt}, they described GPT-NeoX-20B’s architecture and training and evaluated its performance on a range of language-understanding, mathematics, and knowledge-based tasks. They found that GPT-NeoX-20B is a particularly powerful few-shot reasoner and gains far more in performance when evaluated five-shot than similarly sized GPT-3 and FairSeq models.

\subsection{BLOOM}

The proposal introduces BLOOM \cite{workshop2022bloom}, a language model characterized by its expansive scale, featuring 176 billion parameters. This model is a product of the collaborative efforts within BigScience, a consortium comprised of hundreds of researchers. Notably, BLOOM operates as a decoder-only Transformer language model, representing a distinctive architectural choice. The training process involved the utilization of the ROOTS corpus, a comprehensive dataset encompassing content from hundreds of sources \cite{lhoest2021datasets}\cite{laurenccon2022bigscience}. This corpus spans 46 natural languages and 13 programming languages, resulting in a total of 59 languages covered. The deliberate design and training choices made in the development of BLOOM underscore its potential significance in advancing the capabilities of language models, with implications for diverse linguistic contexts. 


This paper \cite{workshop2022bloom} documents the evolution of BLOOM, tracing its development from the inception of its training dataset ROOTS to the formulation of its architecture and tokenizer. The authors delve into the evaluation outcomes of BLOOM and other expansive language models, revealing its competitive performance that undergoes enhancement through multitask finetuning. The authors express the anticipation that the introduction of such a potent multilingual language model will pave the way for novel applications and avenues of research within the domain of large language models.

\subsection{Google’s Generative AI}
\subsubsection{PaLM}
The Pathways Language Model (PaLM) stands as LLM created by Google AI \cite{chowdhery2022palm}. Trained on an extensive dataset encompassing both text and code, PaLM exhibits versatility in its capabilities, demonstrating proficiency across a diverse array of tasks, which include but are not limited to Translation, Summarization, Question Answering, Code generation, and Creative writing.

PaLM employs a transformer-based design, leveraging transformers—a distinct class of neural networks well-suited for natural language processing tasks. These models stand out for their proficiency in acquiring the capability to grasp and understand the nuanced relationships among words and phrases within a given sentence. PaLM, an extraordinary language model, undergoes training on an unprecedented dataset featuring a staggering 540 billion parameters, marking it as the largest dataset ever utilized for language model training. The model showcases remarkable versatility, demonstrating proficiency across diverse tasks such as translation, summarization, question answering, code generation, and even creative writing.
This expansive skill set has facilitated the creation of innovative applications, including a tool designed to assist individuals with dyslexia in enhancing their reading experience. Despite being in the developmental stage, PaLM has already exhibited substantial promise. As development continues, the model is poised to evolve into an even more potent and capable tool, potentially revolutionizing various aspects of our lives.
PaLM's potential extends to enhancing the accuracy and efficiency of numerous tasks, ranging from translation and summarization to question answering. Its versatility also opens doors to the creation of novel applications, further exemplified by its utility in aiding individuals with dyslexia.

Google contends that PaLM 2 demonstrates enhanced reasoning capabilities compared to GPT-4 across diverse benchmarks, with particularly notable advancements observed in tasks such as WinoGrande and DROP. In these specific assessments, PaLM 2 surpasses GPT-4 by a modest margin. However, an examination of the ARC-C benchmark reveals that, although PaLM 2 exhibits some progress, it is not as pronounced as in other benchmark scenarios.
Beyond its improved reasoning, PaLM 2 showcases augmented mathematical prowess, elucidated in a comprehensive 91-page research paper by Google. It is essential to acknowledge the inherent challenges in directly comparing PaLM 2 with GPT-4, owing to the distinct presentation methodologies employed by Google and OpenAI in reporting their respective test results.
Google has deliberately chosen specific comparisons, potentially stemming from instances where PaLM 2 did not perform as effectively as GPT-4. A case in point is the MMLU benchmark, where GPT-4 achieved a score of 86.4, whereas PaLM 2 attained a slightly lower score of 81.2. Similarly, in the HellaSwag benchmark, GPT-4 garnered a score of 95.3, while PaLM 2 achieved a more modest score of 86.8. Lastly, in the ARC-E benchmark, GPT-4 and PaLM 2 secured scores of 96.3 and 89.7, respectively.
These findings underscore the nuanced distinctions in performance between the two models across a spectrum of evaluation tasks.

\subsubsection{Google Bard}

Google Bard constitutes a chatbot founded on the Large Language Model (LLM) PaLM 2. The present iteration of Bard is accessible in English, Japanese, and Korean, with users able to engage with it either via the Google Bard website or through the Google Assistant platform \cite{aydin2023google}.

Key Features of Google Bard:
\begin{itemize}

\item Bidirectional Context Understanding: A pivotal characteristic of BARD lies in its bidirectional methodology, enabling the model to process information from both preceding and succeeding tokens. This key feature enhances the model's capacity to comprehend intricate sentence patterns, resulting in more precise and relevant responses. It is noteworthy that, in contrast, ChatGPT-4 remains unidirectional.

\item Enhanced Language Comprehension: Bard exhibits heightened language comprehension capabilities by considering the complete phrase structure in both forward and backward directions. Similar to ChatGPT-4, it excels in activities such as text completion, question-answering, and a comprehensive understanding of the contextual nuances within a given language input.

\item Language Task Versatility: The bidirectional nature of BARD imparts versatility, rendering it apt for a diverse range of language tasks. Particularly effective in natural language processing applications, BARD excels in tasks where contextual understanding plays a pivotal role, making it a valuable asset in scenarios demanding nuanced language comprehension.
\end{itemize}

This investigation \cite{koga2023evaluating} delves into the efficacy of LLMs, specifically ChatGPT and Google Bard, in predicting neuropathologic diagnoses based on clinical summaries. An analysis was conducted on 25 cases of neurodegenerative disorders presented at Mayo Clinic's brain bank Clinico-Pathological Conferences. The LLMs generated multiple pathologic diagnoses along with their respective rationales, which were then juxtaposed with the conclusive clinical diagnoses provided by physicians.
ChatGPT-3.5, ChatGPT-4, and Google Bard accurately identified primary diagnoses in 32\%, 52\%, and 40\% of cases, respectively. Moreover, correct diagnoses were encompassed in 76\%, 84\%, and 76\% of cases, respectively. These results underscore the potential of artificial intelligence tools such as ChatGPT and Bard in the realm of neuropathology.

\section{Research Challenges and Open Problems}
Despite the remarkable capabilities exhibited by LLMs in both industrial applications and academic endeavors, they are not immune to certain limitations and persistent challenges.

\subsection{Research Challenges}

While LLM models have proven immensely beneficial in advancing scientific discoveries, it is vital to acknowledge and confront the research challenges that have emerged due to their implementation. By acknowledging and actively addressing these challenges, we can ensure continued progress and refinement in this field.

\subsubsection{Computational Costs and Power}
The LLM model is an advanced and intricate AI language model that necessitates significant computational resources for optimal performance. Operating this model can be costly and may demand access to specialized hardware and software systems. Utilizing low-end hardware or systems with limited computational power can lead to slower processing times, decreased accuracy, and various performance-related challenges. Therefore, organizations should thoroughly evaluate their computational resources and capabilities when considering the implementation of LLM models.

\subsubsection{Number of Parameters:}
One of the challenges associated with LLM models is the sheer number of parameters they contain. These models often have an extremely high parameter count, which contributes to their complexity and computational requirements. Managing and training models with a large number of parameters can be resource-intensive, requiring significant computational power and memory. It also increases the risk of over-fitting and can make the model more susceptible to errors or biases present in the training data. Balancing the parameter count to optimize performance and efficiency without compromising model quality is an ongoing challenge in the development and deployment of LLM models.

\subsubsection{Up-to-Date:}
Maintaining up-to-date LLM models necessitates regular retraining or fine-tuning with the most recent data, a process that demands significant resources. Failing to address the matter of outdated training data can result in the generation of misleading or inaccurate information by the models. Therefore, it is crucial to allocate sufficient resources to ensure the timely and comprehensive updating of LLM models to uphold their reliability and relevance.

\subsubsection{Cost:}
The training and utilization of LLMs necessitate significant computational power, involving specialized hardware and substantial time investments for both training and inference. These resources come with associated costs, rendering LLMs inaccessible to individuals or organizations with limited resources.

Furthermore, fine-tuning or customizing LLMs for specific tasks often demands extensive data annotation efforts, which further contribute to the overall cost. The high computational and financial requirements associated with LLMs can create barriers to entry, limiting the widespread adoption and usage of these models, particularly for smaller-scale projects or applications with constrained budgets.

\subsubsection{Data:}
One of the primary difficulties associated with LLM models relates to their extensive access to data, which frequently encompasses personal details like names, addresses, and phone numbers. Although most solutions do not expressly store or disclose this information, there remains a possibility of unintentionally including sensitive data within the input or output of a conversation. As an extensively trained language model with vast amounts of data, LLM models provide users with a formidable capability to generate text, pose relevant questions, and engage in conversations. Nonetheless, employing such models also presents substantial challenges regarding data privacy that demand attention and resolution.

The issue of data sufficiency poses a significant challenge in the realm of LLMs. Despite their impressive capabilities, LLMs heavily rely on the vast amounts of data they are trained on, and their performance is contingent upon the quality and diversity of that data. In scenarios where the training data is insufficient or lacks a comprehensive representation of various domains, the LLMs may struggle to generate accurate and contextually relevant outputs. This data sufficiency issue becomes particularly pronounced when confronted with specialized or niche domains where the availability of training data is limited. 

Data bias in models like ChatGPT refers to the presence of unfair or systematic preferences in the training data, originating from various sources including data collection, societal biases, and inherent biases in the data sources. Language models like ChatGPT are trained on vast amounts of internet text data, which can introduce biases related to gender, race, religion, and other aspects of identity, leading the model to reproduce such biases in its responses. Data bias can result in discriminatory responses, assumptions, or inaccuracies based on the biased patterns learned during training. Addressing data bias is an active research area, aiming to identify and mitigate biases during training to enhance fairness and inclusivity in these models, but it remains a complex challenge requiring continuous efforts and collaboration for meaningful progress. \cite{gill2023chatgpt}.

\subsection{Open Problems}

Language models like ChatGPT, including Large Language Models (LLMs), have limitations despite their impressive performance. These limitations stem from factors such as biases in training data, a lack of common-sense reasoning, and a reliance on statistical patterns. LLMs can produce factually incorrect outputs and struggle with complex prompts, making contextual understanding and misinformation challenges. Ongoing research is essential to improve their accuracy, ensuring more reliable and trustworthy results in various applications and mitigating these limitations.

Future research holds promise in the development of specialized datasets that extend beyond simple data collection from existing sources. Instead, these datasets should be carefully crafted to meet specific requirements, audiences, or problem domains, guided by thoughtful consideration. For instance, in the legal field, constructing a comprehensive dataset would necessitate a deep understanding of legal history and previous cases, serving as a valuable resource for training AI models in legal tasks. Creating such datasets requires meticulous planning in data collection methods to represent real-world complexities accurately, and techniques like data augmentation, transfer learning, and fine-tuning should enhance diversity and quality. Domain-specific datasets tailored to unique needs will be pivotal in achieving high-performance AI models for practical applications and advancing solutions to real-world challenges.

Meanwhile, for future work, it is also essential to ask the question of who the target audience of the model is. Identifying the specific audience categories, such as medical, law, education, etc., will help determine the necessary adaptations and additions to the model to meet their unique requirements. For instance, if the model is intended for the medical field, it should be tailored to handle medical terminology, understand complex medical concepts, and ensure trustworthy results for diagnostic purposes.
As a state-of-the-art review, considering future work is imperative. Moving beyond simple word prediction, the next level of development involves incorporating reasoning capabilities and applying the model to knowledge graphs using APIs. This extension would allow the model to exhibit advanced understanding and reasoning abilities, making it more adept at handling complex and nuanced tasks.
In the context of education, exploring the potential of using the model as a teaching tool opens new avenues for research and development. To make it more applicable in the education field, necessary changes or additions should be considered, such as integrating pedagogical elements, adapting to diverse learning styles, and fostering educational engagement.

While ChatGPT represents a groundbreaking advancement, it is not recommended to seek medical advice from it. However, there is a growing interest in determining when artificial intelligence (AI) will be capable of providing such assistance and how much more accurate LLMs need to become to achieve that level of reliability. These questions are currently being explored by researchers. The continuous advancement of LLMs holds tremendous potential for their future development and application. Ongoing research and innovation are focusing on improving LLMs' contextual understanding, enhancing their reasoning capabilities, and reducing biases. When applied to healthcare, building trust in the model's results becomes paramount. Trustworthiness is crucial in medical diagnosis and treatment recommendations. Ensuring high accuracy, transparency, and validation mechanisms would be essential to gain the trust of medical professionals and patients alike.

For the legal domain, it is crucial to address concerns related to privacy, data protection, and adherence to legal regulations. Additionally, ensuring that the model interprets legal language accurately and provides reliable legal information would be imperative.
In summary, future work involves carefully tailoring the model to specific audience categories, incorporating advanced reasoning capabilities, exploring educational applications, ensuring trustworthiness in healthcare applications, and addressing legal concerns. Addressing these aspects will contribute to the robustness and applicability of the model across various domains, making it more versatile and reliable in practical real-world scenarios.

\subsection{Future Research Works}
\textbf{Autonomous models that generate training data}: The development of autonomous models capable of generating their own training data marks a significant step towards self-sufficiency in LLMs. Current LLMs rely heavily on curated datasets to learn and generalize patterns. However, this approach is often limited by the availability and quality of training data. Future research is focusing on creating models that can autonomously generate synthetic data, enabling them to continuously adapt and improve their performance without requiring additional human-labeled datasets.
These autonomous models can simulate realistic scenarios, introduce variations, and even create edge cases that may be challenging for current models. By doing so, they not only enhance their own training but also become more robust in handling unforeseen situations. This research path opens up possibilities for more efficient and adaptive LLMs, especially in domains where collecting large and diverse datasets is challenging or costly.

\textbf{Models that can validate their information}: The ability of models to validate their information is crucial for ensuring the reliability and trustworthiness of LLMs. In the current LLM, models are trained on historical data and then deployed in real-world scenarios. However, these models may encounter situations or data distributions that differ significantly from their training data, leading to potential inaccuracies or biases.
Future research is focused on developing models with built-in mechanisms for self-validation. These models can assess the quality and relevance of the data they receive during inference, flagging potential issues and uncertainties. This self-awareness enables models to provide more accurate and reliable predictions while also identifying when they may be out of their depth. This research path is crucial for LLMs in sensitive domains such as healthcare, finance, and autonomous vehicles, where the consequences of incorrect results can be severe.

\textbf{Rise of Sparse Expert Models}: Sparse expert models represent a departure from the traditional approach of large, dense neural networks. Instead of relying on massive numbers of parameters, sparse expert models focus on a select set of specialized neurons or modules that excel at specific tasks. This approach is inspired by the human brain, where different regions specialize in different cognitive functions.
Research in this area aims to create more efficient and interpretable LLMs. Sparse expert models can be particularly beneficial in scenarios where computational resources are limited, such as edge devices or real-time applications. Additionally, the sparsity of these models often makes them more interpretable, addressing concerns about the "black box" nature of LLMs. The rise of sparse expert models signifies a shift towards more resource-efficient and interpretable LLM solutions, paving the way for their application in a wider range of real-world settings.

\section{Conclusion}
This survey undertakes a comprehensive examination of recent advancements in Large Language Models (LLMs), offering a detailed overview of prominent models in both industrial and academic domains. A thorough analysis of these LLMs was conducted, and attempts were made to compare and evaluate them based on available evidence. It is noteworthy that not all solutions yielded publicly available evaluation results. Despite this limitation, the survey endeavors to encompass the latest literature and solutions about LLMs, serving as a valuable reference resource for researchers and engineers interested in staying abreast of developments in this dynamic field.


\section*{Authors}

\begin{tabular}{p{3cm}p{12cm}}
        \includegraphics[valign=t,width=2.8cm]{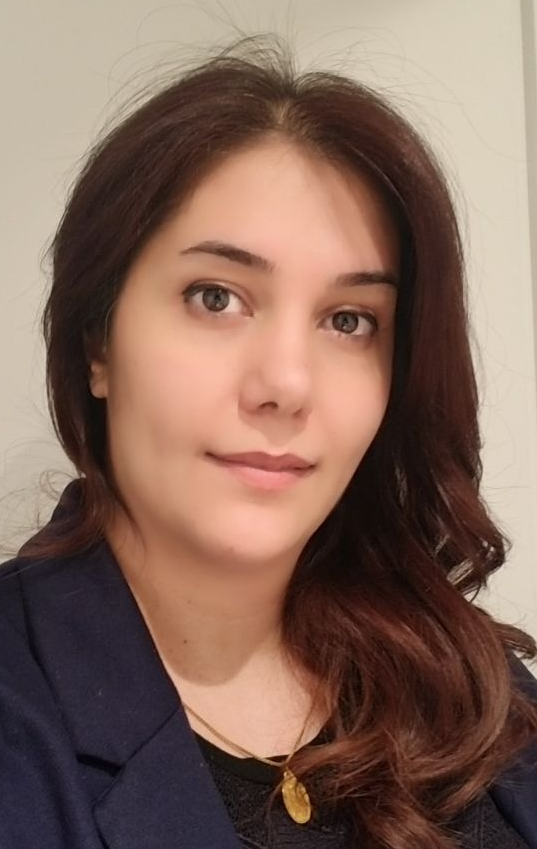} \rotatebox[origin=l]{0}{Hanieh Alipour} &  Hanieh Alipour is a Senior Data Scientist of the Industry and Customer Experience Data Science Research team. Her force is in the field of artificial intelligence and natural language processing. Currently, she is working on Large Language Models (LLM) for different use cases such as email assistance. Dr. Hanieh Alipour earned her Ph.D. in Computer Science from Concordia University Montreal. Her research focus delved deep into the realm of Model-driven, Cloud Auto-scaling service, and Machine Learning. \\
        \includegraphics[valign=t,width=2.8cm]{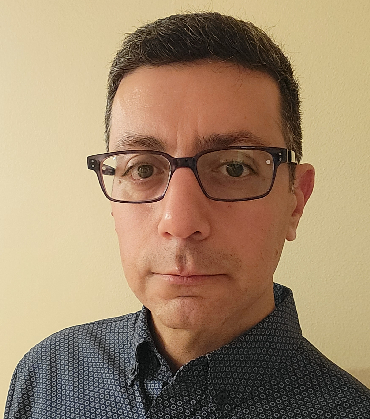} \rotatebox[origin=l]{0}{Nick Pendar} & Nick Pendar is the Chief AI Expert at SAP Customer Experience where he helps build AI services with tangible business value. He has a Ph.D. in computational linguistics from the University of Toronto and has two decades of academic and industry experience in the fields of natural language processing and machine learning. Nick was a founding member of the IEEE International Conference on Semantic Computing and has served as an AI advisor on multiple startups.\\
        \includegraphics[valign=t,width=2.8cm]{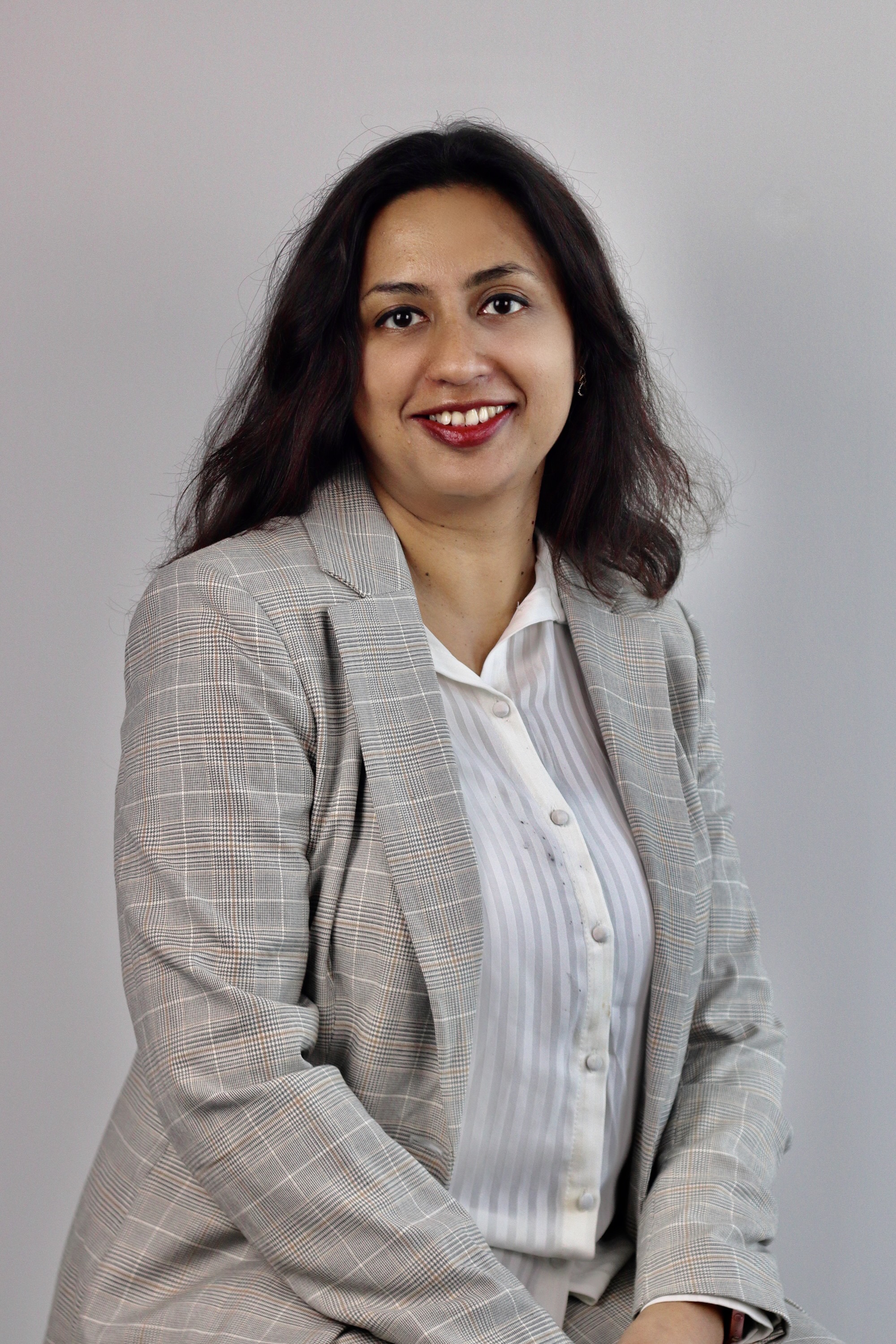} \rotatebox[origin=l]{0}{Kohinoor Roy} & Kohinoor Roy is a Senior Data Scientist of the Industry and Customer Experience Data Science Research team. She has worked in different machine learning and deep learning algorithms of Commerce and NLP-based email assistants. Recently, she has worked on personalized recommendation algorithms, image recognition, and segmentation-based styling products and home decor items. She has been working in SAP as a Data Scientist for more than 6 years now and before that, she has 10 years of software engineering experience in different roles and domains.
\end{tabular}

\vspace{2cm}

\begin{thebibliography}{4}

\bibitem{y2022large}
  y Arcas, Blaise Ag{\"u}era, Do large language models understand us?, Daedalus, Volume 151,  Number 2, Pages 183-197, MIT Press One Rogers Street, Cambridge, MA 02142-1209, USA, 2022.


\bibitem{hauser2002faculty}
Hauser, Marc D and Chomsky, Noam and Fitch, W Tecumseh, The faculty of language: what is it, who has it, and how did it evolve?,
  Volume 298,
  Number 5598,
  Pages 1569-1579,American Association for the Advancement of Science, 2002.

\bibitem{rosenfeld2000two}
Rosenfeld, Ronald, Two decades of statistical language modeling: Where do we go from here?,
Proceedings of the IEEE, 
  Volume 88,
  Number 8,
  Pages 1270-1278, 2000.

\bibitem{wang2019superglue}
Wang, Alex and Pruksachatkun, Yada and Nangia, Nikita and Singh, Amanpreet and Michael, Julian and Hill, Felix and Levy, Omer and Bowman, Samuel, Superglue: A stickier benchmark for general-purpose language understanding systems, 
Advances in neural information processing systems,
  Volume 32, 2019.

\bibitem{adiwardana2020towards}
Adiwardana, Daniel and Luong, Minh-Thang and So, David R and Hall, Jamie and Fiedel, Noah and Thoppilan, Romal and Yang, Zi and Kulshreshtha, Apoorv and Nemade, Gaurav and Lu, Yifeng and others, Towards a human-like open-domain chatbot,
arXiv:2001.09977, 2020.

\bibitem{GPT4}
OpenAI, GPT-4 Technical Report,
arXiv:2303.08774, 2023.

\bibitem{mu2023navigating}
Mu, Yida and Wu, Ben P and Thorne, William and Robinson, Ambrose and Aletras, Nikolaos and Scarton, Carolina and Bontcheva, Kalina and Song, Xingyi, 
Navigating Prompt Complexity for Zero-Shot Classification: A Study of Large Language Models in Computational Social Science,
  arXiv:2305.14310, 2023.

\bibitem{kopf2023openassistant}
K{\"o}pf, Andreas and Kilcher, Yannic and von R{\"u}tte, Dimitri and Anagnostidis, Sotiris and Tam, Zhi-Rui and Stevens, Keith and Barhoum, Abdullah and Duc, Nguyen Minh and Stanley, Oliver and Nagyfi, Rich{\'a}rd and others, OpenAssistant Conversations--Democratizing Large Language Model Alignment,
 arXiv:2304.07327, 2023.

\bibitem{zhao2021calibrate}
Zhao, Zihao and Wallace, Eric and Feng, Shi and Klein, Dan and Singh, Sameer, Calibrate before use: Improving few-shot performance of language models,
International Conference on Machine Learning,
12697-12706, 2021.

\bibitem{zhou2022large}
Zhou, Yongchao and Muresanu, Andrei Ioan and Han, Ziwen and Paster, Keiran and Pitis, Silviu and Chan, Harris and Ba, Jimmy, Large language models are human-level prompt engineers,
arXiv:2211.01910, 2022.


\bibitem{kojima2022large}
Kojima, Takeshi and Gu, Shixiang Shane and Reid, Machel and Matsuo, Yutaka and Iwasawa, Yusuke, Large language models are zero-shot reasoners,
Advances in neural information processing systems,
  Volume 35, Pages 22199-22213, 2022.

\bibitem{prystawski2023think}
Prystawski, Ben and Goodman, Noah D, Why think step-by-step? Reasoning emerges from the locality of experience,
 arXiv:2304.03843, 2023.

\bibitem{vaswani2017attention}
Vaswani, Ashish and Shazeer, Noam and Parmar, Niki and Uszkoreit, Jakob and Jones, Llion and Gomez, Aidan N and Kaiser, {\L}ukasz and Polosukhin, Illia, Attention is all you need,
Advances in neural information processing systems,
  Volume 30, 2017.

\bibitem{zhang2019root}
Zhang, Biao and Sennrich, Rico, mean square layer normalization,
Advances in Neural Information Processing Systems,
  Volume 32, 2019.

\bibitem{shazeer2020glu}
Shazeer, Noam, Glu variants improve transformer,
arXiv:2002.05202, 2020.

\bibitem{su2021roformer}
Su, Jianlin and Lu, Yu and Pan, Shengfeng and Murtadha, Ahmed and Wen, Bo and Liu, Yunfeng,
Roformer: Enhanced transformer with rotary position embedding,
arXiv:2104.09864, 2021.

\bibitem{touvron2302llama}
Touvron, Hugo and Lavril, Thibaut and Izacard, Gautier and Martinet, Xavier and Lachaux, Marie-Anne and Lacroix, Timoth{\'e}e and Rozi{\`e}re, Baptiste and Goyal, Naman and Hambro, Eric and Azhar, Faisal and others, LLaMA: open and efficient foundation language models, 2023, https://arxiv. org/abs/2302.13971.


\bibitem{brown2020language}
Brown, Tom and Mann, Benjamin and Ryder, Nick and Subbiah, Melanie and Kaplan, Jared D and Dhariwal, Prafulla and Neelakantan, Arvind and Shyam, Pranav and Sastry, Girish and Askell, Amanda and others, Language models are few-shot learners,
Advances in neural information processing systems,
  Volume 33,
  Pages 1877-1901, 2020.


\bibitem{rae2021scaling}
Rae, Jack W and Borgeaud, Sebastian and Cai, Trevor and Millican, Katie and Hoffmann, Jordan and Song, Francis and Aslanides, John and Henderson, Sarah and Ring, Roman and Young, Susannah and others, Scaling language models: Methods, analysis \& insights from training gopher, arXiv:2112.11446, 2021.

\bibitem{hoffmann2022training}
Hoffmann, Jordan and Borgeaud, Sebastian and Mensch, Arthur and Buchatskaya, Elena and Cai, Trevor and Rutherford, Eliza and Casas, Diego de Las and Hendricks, Lisa Anne and Welbl, Johannes and Clark, Aidan and others, Training compute-optimal large language models, arXiv:2203.15556, 2022.

\bibitem{chowdhery2022palm}
Chowdhery, Aakanksha and Narang, Sharan and Devlin, Jacob and Bosma, Maarten and Mishra, Gaurav and Roberts, Adam and Barham, Paul and Chung, Hyung Won and Sutton, Charles and Gehrmann, Sebastian and others, Palm: Scaling language modeling with pathways,
arXiv:2204.02311, 2022.

\bibitem{zhang2022opt}
Zhang, Susan and Roller, Stephen and Goyal, Naman and Artetxe, Mikel and Chen, Moya and Chen, Shuohui and Dewan, Christopher and Diab, Mona and Li, Xian and Lin, Xi Victoria and others,  Opt: Open pre-trained transformer language models,
arXiv:2205.01068, 2022.

\bibitem{wang2021gpt}
Wang, Ben and Komatsuzaki, Aran, GPT-J-6B: A 6 billion parameter autoregressive language model, 2021.

\bibitem{black2022gpt}
Black, Sid and Biderman, Stella and Hallahan, Eric and Anthony, Quentin and Gao, Leo and Golding, Laurence and He, Horace and Leahy, Connor and McDonell, Kyle and Phang, Jason and others. Gpt-neox-20b: An open-source autoregressive language model,
arXiv:2204.06745, 2022.

\bibitem{roumeliotis2023llama}
Roumeliotis, Konstantinos I and Tselikas, Nikolaos D and Nasiopoulos, Dimitrios K, Llama 2: Early Adopters' Utilization of Meta's New Open-Source Pretrained Model,
2023.

\bibitem{taori2023alpaca}
Taori, Rohan and Gulrajani, Ishaan and Zhang, Tianyi and Dubois, Yann and Li, Xuechen and Guestrin, Carlos and Liang, Percy and Hashimoto, Tatsunori B, Alpaca: A strong, replicable instruction-following model, 
Stanford Center for Research on Foundation Models. https://crfm. stanford. edu/2023/03/13/alpaca. html,
  Volume 3,
  Number 6,
  Pages 7, 2023.

\bibitem{workshop2022bloom}
Workshop, BigScience and Scao, Teven Le and Fan, Angela and Akiki, Christopher and Pavlick, Ellie and Ili{\'c}, Suzana and Hesslow, Daniel and Castagn{\'e}, Roman and Luccioni, Alexandra Sasha and Yvon, Fran{\c{c}}ois and others,
Bloom: A 176b-parameter open-access multilingual language model,
arXiv:2211.05100, 2022.

\bibitem{kang2020natural}
Kang, Yue and Cai, Zhao and Tan, Chee-Wee and Huang, Qian and Liu, Hefu, Natural language processing (NLP) in management research: A literature review, Journal of Management Analytics,
  Volume 7,
  Number 2,
  Pages 139-172, 2020.

\bibitem{nadkarni2011natural}
Nadkarni, Prakash M and Ohno-Machado, Lucila and Chapman, Wendy W,
Natural language processing: an introduction, Journal of the American Medical Informatics Association,
  Volume 18,
  Number 5,
  pages=544-551, 2011.

\bibitem{nallapati2004discriminative}
Nallapati, Ramesh, Discriminative models for information retrieval,
Proceedings of the 27th annual international ACM SIGIR conference on Research and development in information retrieval,
  Pages 64-71, 2004.

\bibitem{he2020kgplm}
He, Bin and Jiang, Xin and Xiao, Jinghui and Liu, Qun, Kgplm: Knowledge-guided language model pre-training via generative and discriminative learning, arXiv:2012.03551, 2020.


\bibitem{zhu2022generative}
Zhu, Qihao and Luo, Jianxi, Generative pre-trained transformer for design concept generation: an exploration, Proceedings of the design society,
  Volume 2,
  Pages 1825-1834, 2022.

\bibitem{ciosici2022training}
Ciosici, Manuel R and Derczynski, Leon, Training a T5 Using Lab-sized Resources,
arXiv:2208.12097, 2022.

\bibitem{thapa2023chatgpt}
Thapa, Surendrabikram and Adhikari, Surabhi, ChatGPT, bard, and large language models for biomedical research: opportunities and pitfalls, Springer, 
  Pages 1-5, 2023.

\bibitem{lim2023benchmarking}
Lim, Zhi Wei and Pushpanathan, Krithi and Yew, Samantha Min Er and Lai, Yien and Sun, Chen-Hsin and Lam, Janice Sing Harn and Chen, David Ziyou and Goh, Jocelyn Hui Lin and Tan, Marcus Chun Jin and Sheng, Bin and others, Benchmarking large language models’ performances for myopia care: a comparative analysis of ChatGPT-3.5, ChatGPT-4.0, and Google Bard, Elsevier 2023.

\bibitem{koga2023evaluating}
Koga, Shunsuke and Martin, Nicholas B and Dickson, Dennis W, Evaluating the performance of large language models: ChatGPT and Google Bard in generating differential diagnoses in clinicopathological conferences of neurodegenerative disorders, Brain Pathology (Wiley Online Library), 2023.

\bibitem{casper2023open}
Casper, Stephen and Davies, Xander and Shi, Claudia and Gilbert, Thomas Krendl and Scheurer, J{\'e}r{\'e}my and Rando, Javier and Freedman, Rachel and Korbak, Tomasz and Lindner, David and Freire, Pedro and others, Open problems and fundamental limitations of reinforcement learning from human feedback, arXiv:2307.15217, 2023.

\bibitem{wei2022chain}
Wei, Jason and Wang, Xuezhi and Schuurmans, Dale and Bosma, Maarten and Xia, Fei and Chi, Ed and Le, Quoc V and Zhou, Denny and others, Chain-of-thought prompting elicits reasoning in large language models, Advances in Neural Information Processing Systems,
  Volume 35,
  Pages 24824-24837, 2022.

\bibitem{openAI}
https://openai.com/about.


\bibitem{gill2023chatgpt}
Gill, Sukhpal Singh and Kaur, Rupinder, ChatGPT: Vision and challenges,
Internet of Things and Cyber-Physical Systems,
  Volume 3, Pages 262-271, Elsevier 2023.

\bibitem{ray2023chatgpt}
Ray, Partha Pratim, ChatGPT: A comprehensive review on background, applications, key challenges, bias, ethics, limitations and future scope, 
Internet of Things and Cyber-Physical Systems, Elsevier 2023.

\bibitem{gao2020pile}
Gao, Leo and Biderman, Stella and Black, Sid and Golding, Laurence and Hoppe, Travis and Foster, Charles and Phang, Jason and He, Horace and Thite, Anish and Nabeshima, Noa and others, The pile: An 800gb dataset of diverse text for language modeling, arXiv:2101.00027,2020.

\bibitem{nguyen2019transformers}
Nguyen, Toan Q and Salazar, Julian, Transformers without tears: Improving the normalization of self-attention, arXiv:1910.05895, 2019.

\bibitem{Wang}
Ben Wang, Mesh-Transformer-JAX: Modelparallel implementation of transformer language model with JAX, 2021.

\bibitem{koubaa2023gpt}
Koubaa, Anis, GPT-4 vs. GPT-3.5: A concise showdown, Preprints,2023.

\bibitem{chatgpt4}
OpenAI, GPT-4 Technical Report,2023.

\bibitem{aydin2023google}
AYDIN, {\"O}mer, Google Bard generated literature review: metaverse,
  Journal of AI,
  Volume 7,
  Number 1,
  Pages 1-14, {\.I}zmir Academy Association,2023.

\bibitem{lhoest2021datasets}
Lhoest, Q and del Moral, AV and Jernite, Y and Thakur, A and von Platen, P and Patil, S and Chaumond, J and Drame, M and Plu, J and Tunstall, L and others, Datasets: a community library for natural language processing, arXiv preprint arXiv:2109.02846,2021.

\bibitem{laurenccon2022bigscience}
Lauren{\c{c}}on, Hugo and Saulnier, Lucile and Wang, Thomas and Akiki, Christopher and Villanova del Moral, Albert and Le Scao, Teven and Von Werra, Leandro and Mou, Chenghao and Gonz{\'a}lez Ponferrada, Eduardo and Nguyen, Huu and others, The bigscience roots corpus: A 1.6 tb composite multilingual dataset, Advances in Neural Information Processing Systems,
  Volume 35, Pages 31809-31826, 2022.

\bibitem{kasneci2023chatgpt}
Kasneci, Enkelejda and Se{\ss}ler, Kathrin and K{\"u}chemann, Stefan and Bannert, Maria and Dementieva, Daryna and Fischer, Frank and Gasser, Urs and Groh, Georg and G{\"u}nnemann, Stephan and H{\"u}llermeier, Eyke and others, ChatGPT for good? On opportunities and challenges of large language models for education, Learning and individual differences (Elsevier), volume 103, pages 102274,2023.

\bibitem{liu2023summary}
Liu, Yiheng and Han, Tianle and Ma, Siyuan and Zhang, Jiayue and Yang, Yuanyuan and Tian, Jiaming and He, Hao and Li, Antong and He, Mengshen and Liu, Zhengliang and others, Summary of chatgpt-related research and perspective towards the future of large language models, Meta-Radiology(Elsevier), pages 100017, 2023.

\bibitem{zhu2023large}
Zhu, Yutao and Yuan, Huaying and Wang, Shuting and Liu, Jiongnan and Liu, Wenhan and Deng, Chenlong and Dou, Zhicheng and Wen, Ji-Rong, Large language models for information retrieval: A survey, arXiv:2308.07107, 2023.

\bibitem{wang2023aligning}
Wang, Yufei and Zhong, Wanjun and Li, Liangyou and Mi, Fei and Zeng, Xingshan and Huang, Wenyong and Shang, Lifeng and Jiang, Xin and Liu, Qun, Aligning large language models with human: A survey, arXiv:2307.12966, 2023.

\bibitem{min2023recent}
Min, Bonan and Ross, Hayley and Sulem, Elior and Veyseh, Amir Pouran Ben and Nguyen, Thien Huu and Sainz, Oscar and Agirre, Eneko and Heintz, Ilana and Roth, Dan, Recent advances in natural language processing via large pre-trained language models: A survey, ACM Computing Surveys, volume 56, pages 1-40, 2023.

\bibitem{hadi2023survey}
Hadi, Muhammad Usman and Qureshi, Rizwan and Shah, Abbas and Irfan, Muhammad and Zafar, Anas and Shaikh, Muhammad Bilal and Akhtar, Naveed and Wu, Jia and Mirjalili, Seyedali and others, A survey on large language models: Applications, challenges, limitations, and practical usage,
Authorea Preprints, 2023.

\end{thebibliography}
\end{document}